\documentclass[runningheads]{llncs}

\usepackage{eccv}

\usepackage{eccvabbrv}
\usepackage{graphicx}
\usepackage{booktabs}
\usepackage[accsupp]{axessibility}

\usepackage{color}
\usepackage{multirow}
\usepackage{algorithm}
\usepackage{algorithmicx}
\usepackage{algpseudocode}
\usepackage{wrapfig}
\usepackage{caption}
\usepackage{tikz}
\usetikzlibrary{shapes.geometric,arrows.meta,positioning,calc}
\usepackage{pgfplots}
\pgfplotsset{compat=1.18}

\definecolor{svblue}{RGB}{70, 130, 180}      
\definecolor{mvorange}{RGB}{230, 126, 34}     
\definecolor{mvgold}{RGB}{243, 156, 18}       

\definecolor{tentative}{RGB}{149, 165, 166}   
\definecolor{active}{RGB}{46, 204, 113}       
\definecolor{quasiactive}{RGB}{52, 152, 219}  
\definecolor{inactive}{RGB}{241, 196, 15}     
\definecolor{terminated}{RGB}{231, 76, 60}    

\definecolor{detection}{RGB}{155, 89, 182}    
\definecolor{communication}{RGB}{26, 188, 156}
\definecolor{fusion}{RGB}{230, 126, 34}       
\definecolor{background}{RGB}{250, 250, 250}  

\definecolor{darkgray}{RGB}{52, 73, 94}       
\definecolor{medgray}{RGB}{127, 140, 141}     
\definecolor{lightgray}{RGB}{189, 195, 199}   

\makeatletter
\ifx\columnwidth\undefined
  \newlength\columnwidth
\fi
\makeatother

\usepackage[pagebackref,breaklinks,colorlinks,citecolor=eccvblue]{hyperref}

\title{Fully Distributed Multi-View 3D Tracking in Real-Time}
\titlerunning{MV3DT}

\author{Byron Hernandez\inst{1,2} \and
Fangyu Li\inst{2} \and
Aotian Wu\inst{2} \and
Paul J.\ Shin\inst{2} \and
Kaustubh Purandare\inst{2} \and
Henry Medeiros\inst{1}}

\authorrunning{B.~Hernandez et al.}

\institute{University of Florida, Gainesville, FL, USA\\
\email{\{bhernandezosorio,hmedeiros\}@ufl.edu}
\and
NVIDIA Corporation, Santa Clara, CA, USA\\
\email{\{bhernandez,fangyul,aotianw,pshin,kpurandare\}@nvidia.com}}

\begin{document}
\maketitle

\begin{abstract}
Multi-camera tracking with overlapping fields of view typically relies on centralized fusion, which creates computational bottlenecks that prevent deployment at scale. We present \textbf{MV3DT}, a fully distributed framework for real-time multi-view 3D tracking that achieves accurate identity propagation and occlusion recovery through peer-to-peer coordination, eliminating the need for central aggregation. 
Each camera node executes a lightweight modular pipeline comprising monocular 3D perception, distributed multi-view association, and collaborative fusion via lightweight messaging. 
MV3DT achieves 96.5\% IDF1, 93.1\% MOTA, and 94.6\% MOTP on WILDTRACK, competitive with state-of-the-art centralized methods, and unprecedented 41.7\% IDF1 and 50.9\% MOTA on SCOUT while demonstrating superior scalability: sustaining 30 FPS on 100 cameras with $<$10ms inter-camera latency and only 2.2\% communication overhead. 
MV3DT operates in a zero-shot regime given camera calibrations, requiring no scene-specific learning and making it directly deployable in new environments. These results establish MV3DT as a practical solution for real-time multi-view tracking in large-scale overlapping camera networks.
\end{abstract}

\section{Introduction} \label{sec:introduction}

\begin{figure}[ht!]
    \centering
    \includegraphics[width=\textwidth]{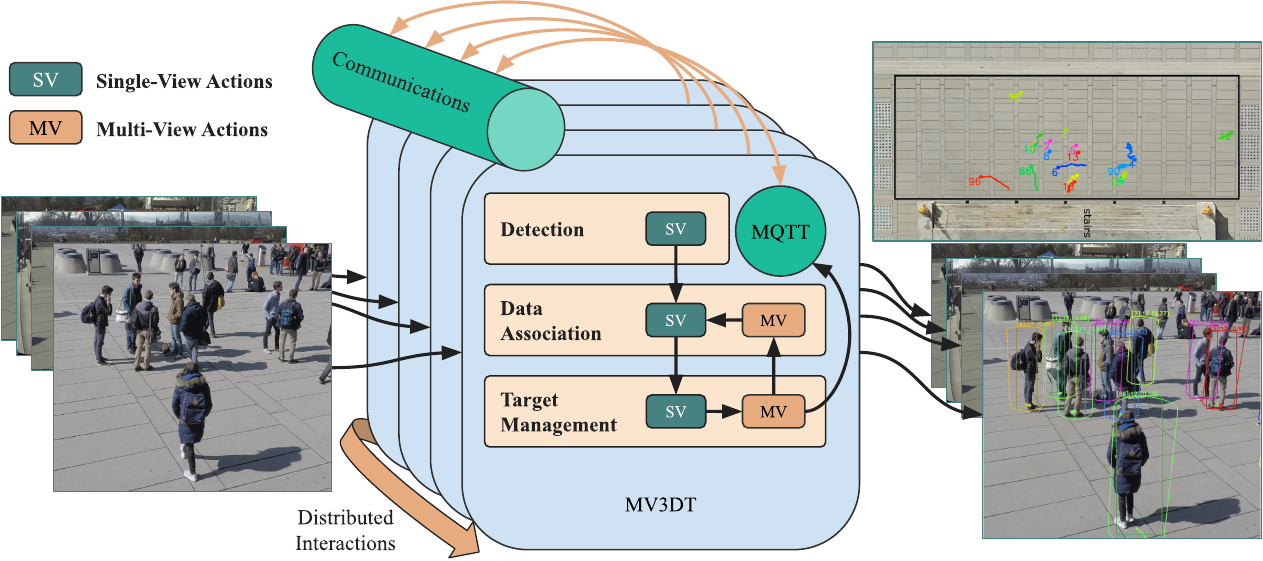}
    \caption{
    \textbf{MV3DT Overview.} 
    MV3DT deploys a modular pipeline on each camera node without requiring a central server. 
    \textbf{Monocular Detection} extracts 2D bounding boxes.
    Then, 3D foot location estimates and full-body bounding boxes are computed for \textbf{Data Association}, where detection-to-targets matches, both intra-view and multi-view, are found using several similarity measures.
    \textbf{Target Management} maintains target state and ID consistency across overlapping cameras through distributed ID propagation, and integrates Kalman filtering for multi-view measurement fusion.
    \textbf{Distributed Communication} uses MQTT publish/subscribe messaging for peer-to-peer coordination. Each camera maintains a local database of shared target states, enabling coordinated tracking without centralized aggregation.
    MV3DT achieves highly effective ID propagation and multi-view integration through \textit{fully-distributed interactions}, allowing for online and real-time deployment of large camera networks.
    }
    \label{fig:general}
\end{figure}

Multi-camera multi-target tracking (MCMT) is a prevalent problem in computer vision. 
Large-scale applications such as warehouse monitoring and intelligent cities require tens to thousands of cameras to effectively cover the region of interest~\cite{naphade2017nvidia}.
The increasing participation in the AI City Challenge reflects the growing importance of large-scale MCMT  ~\cite{naphade2023aicitychallenge,wang2024aicitychallenge,tang2025aicitychallenge}.
MCMT techniques can be classified as centralized, decentralized, or distributed, depending on how they execute processes and aggregate data~\cite{iguernaissi2019people}.
They may also focus on camera topologies with overlapping or non-overlapping fields of view (FOV).
The amount of FOV overlap determines the availability of multi-view geometric cues to improve tracking accuracy.
On the other hand, non-overlapping camera systems typically use appearance representations and trajectory prediction based on camera linking models to achieve effective tracking~\cite{olagoke2020literature, amosa2023multi}.

Centralized approaches leverage overlapping camera setups to exploit global information from all cameras on a \textit{single fusion stage}, which can improve accuracy but concentrates all computation and communication in one logical node~\cite{iguernaissi2019people}. 
In large deployments, such centralized fusion often becomes impractical due to bandwidth, latency, and robustness constraints.
Many distributed MCMT methods target non-overlapping or sparsely overlapping camera topologies, where cross-camera association relies primarily on appearance and temporal constraints.
In these settings, cameras often operate without explicit 3D calibration and maintain consistency mainly through label and appearance exchange.
The current paradigm of distributed techniques for overlapping camera setups relies on parallel single-camera tracking (SCT) processes followed by a centralized multi-camera aggregation stage. 
This \textit{two-stage} dependency on a central entity hinders real-time deployment and limits scalability.

Overlapping FOVs provide complementary 3D geometric information that reduces the impact of occlusions, which is one of the most challenging issues in multi-object tracking~\cite{black2006multi,kim2023addressing}. 
Although centralized methods have long benefited from overlapping FOVs, fully distributed MCMT systems that operate directly in the 3D ground plane and are demonstrated at large scale on fixed, calibrated overlapping camera networks remain scarce~\cite{amosa2023multi, yang2022distributed, previtali2017distributed}.
The fundamental difficulty is a self-conflicting requirement: the multi-view geometric cues improve tracking accuracy, yet exploiting them in a distributed manner is challenging and has limited the scalability of accurate multi-view tracking.

One of the main challenges for scalable deployment is the availability of computing resources. 
Advances in computer and communication technology have enabled larger interconnected camera networks for real-time supervision.
Simultaneously, models and algorithm complexity have grown proportionally. 
Therefore, accurately tracking multiple objects in real-time across large camera networks while leveraging overlapping views remains an open challenge. 
Recent fully distributed approaches already demonstrate that peer-to-peer coordination can sustain online multi-camera tracking~\cite{yang2022distributed, previtali2017distributed}, but they primarily operate in 2D image space and do not exploit calibrated 3D ground-plane geometry in large-scale overlapping camera networks.

As illustrated in~\cref{fig:general}, we propose a fully distributed MCMT framework in which a modular pipeline  processes each video feed in parallel without a central aggregator. 
Each camera node executes a pipeline comprising modules for data association, target management, motion estimation, monocular 3D perception, inter-camera communication, and distributed multi-view fusion. 
The framework enables multi-view identity propagation and 3D fusion through lightweight inter-camera communication, allowing each node to reason locally while achieving globally consistent associations across overlapping views.

Our main contributions are highlighted below:
\begin{itemize}
    \item A fully distributed multi-view 3D tracking framework for calibrated overlapping cameras, in which each node performs local 3D tracking, peer‑to‑peer ID propagation, and multi-view fusion without a central aggregation server.
    \item An occlusion-aware monocular 3D detector with mechanisms that turn noisy single‑view detections into reliable multi‑view measurements.
    \item A three‑stage distributed ID propagation protocol that ensures global ID convergence without a central node.
    \item Large‑scale evaluation with state-of-the-art accuracy on standard benchmarks, measured communication overhead, and synchronized deployment.
\end{itemize}

\section{Related Work} \label{sec:related}
Early object tracking systems based on multiple cameras~\cite{stein1999tracking,lee2000monitoring} used triangulation techniques or exploited the geometry of the scene~\cite{mikic1998video} to combine information from different perspectives. 
One of the main motivations for the development of multi-camera tracking techniques was the resolution of target occlusions in single-view systems~\cite{black2006multi}. 
These approaches were designed to operate on a single computer or on multiple computers orchestrated by a leader.
As the scale of multi-camera systems increased, it became clear that these early \textit{centralized} approaches were limited to small areas covered by few cameras~\cite{taj2011distributed}.

\textit{Decentralized} approaches group cameras into clusters and designate lead nodes or cluster heads to aggregate information from neighboring cameras, reducing communication overhead~\cite{medeiros2008distributed, yoder2010cluster}.
Cluster heads coordinate within their groups and communicate summaries across cluster boundaries.
Fully \textit{distributed} approaches rely on peer-to-peer strategies in which all camera nodes operate as equal participants without hierarchical coordination entities~\cite{chen2011adaptive}.
In such systems, cameras exchange information and reach consensus through decentralized algorithms, eliminating any dependency on leader nodes or coordinators. 
Distributed methods have mainly focused on camera networks with disjoint FOVs so far, where appearance-based object re-identification (reID) and inter-camera linking suffice for cross-camera association.
Recent fully distributed MCMT systems further demonstrate that peer-to-peer coordination can sustain online multi-camera tracking.
Some approaches tackle the association problem by sharing ID labels and appearance features across cameras and maintaining a distributed label--appearance table to reach ID consensus~\cite{yang2022distributed}; others share full tracklets across cameras and fuse these hypotheses into consistent multi-camera trajectories~\cite{previtali2017distributed}.
However, these methods operate primarily in 2D image space and cannot perform 3D ground-plane tracking in large-scale overlapping camera networks.

Camera topology shapes algorithm design~\cite{olagoke2020literature, amosa2023multi}.
For non-overlapping or sparsely overlapping networks, linking models discover spatial or topological connectivity between camera views, learning which cameras observe adjacent or connected regions to establish inter-camera associations~\cite{jiang2018online, quach2021dyglip}.
Recent approaches present reID strategies when targets are not visible for prolonged periods~\cite{ristani2018features, nguyen2023multi}. 
These methods rely on discriminative appearance features, which are also used for multi-camera associations.

For overlapping FOVs, geometric approaches are commonly used for association.
Examples include the projection of image coordinates onto a global coordinate system~\cite{lee2000monitoring} and homography-based matching~\cite{siddique2022tracking}. 
Recent methods leverage transformer-based architectures and bird's-eye view (BEV) representations to aggregate multi-view information early in the pipeline~\cite{teepe2024lifting,wang2024bev}. 
BEVFormer~\cite{li2022bevformer} introduces spatiotemporal transformers that project multi-camera features onto a unified BEV space, enabling robust 3D object detection and tracking. 
Building on this, TrackTacular~\cite{teepe2024lifting} combines temporal feature aggregation with appearance and motion cues for multi-view pedestrian and vehicle tracking. 
BEV-SUSHI~\cite{wang2024bev} extends the BEV paradigm with hierarchical graph neural networks for long-term identity association. 
MVTrajecter~\cite{yamane2025mvtrajecter} incorporates BEV motion and appearance costs, achieving state-of-the-art performance on pedestrian benchmarks. 
Other recent methods explore end-to-end temporal aggregation~\cite{yang2024end} and unified graph-based frameworks, such as the Unified Message Passing Network (UMPN)~\cite{engilberge2025one}.

Tracking accuracy depends on detection quality~\cite{bewley2016sort}, and multi-view detection fusion methods follow the same taxonomy: in non-overlapping networks, each camera runs a 2D detector and reID or linking models handle cross-camera association; in overlapping scenarios, centralized methods may perform fusion at detection time with multi-view or BEV detectors, while distributed systems keep detection local (single-view 2D per node) and perform 3D reasoning at association. Widely used single-view 2D detectors include YOLO~\cite{redmon2016yolo, redmon2017yolo9000, jiang2022yolos}, DETR and its variants~\cite{carion2020detr, zong2023codetrs, zhao2024rtdetr}, and recent YOLO iterations~\cite{wang2024yolov9,wang2024yolov10}. Multi-view detectors~\cite{dong2024mv, daryani2025camuvid, engilberge2023two, aung2024enhancing} improve accuracy in overlapping FOVs but require centralized aggregation and do not scale to distributed deployments.

\section{Multi-Object Tracking Framework} \label{sec:methods}

MV3DT introduces a novel fully distributed and modular MCMT paradigm. 
Our framework aims for a real-time, online, and accurate pipeline for multi-object tracking on multiple cameras; occlusion handling and scalability are key objectives.
Similar in philosophy to~\cite{medeiros2008distributed, wang2013distributed}, we exploit simultaneous information from multiple views in a fully 3D setting.
We use a lightweight peer-to-peer communication strategy to share multi-view information and resolve tracking ambiguities in real-time.
Rather than resorting to a centralized multi-view tracking mechanism or aggregating the results of multiple single-view trackers, our method treats each camera as an independent agent.
Hence, it can be instantiated as a single process that communicates with other cameras within a communication network.
This section describes the core components: detection, data association, target management, multi-view fusion, and communications.

\subsection{Object Detection Module}
This module produces a set of object bounding boxes $\{\mathbf{b}_d\}$ for each input frame, where $\mathbf{b}_d = [u, v, w, h]$, and $(u, v)$ are the pixel coordinates of its top-left corner and $(w, h)$ are its width and height.

\subsubsection{Monocular Foot Localization with Occlusion Handling.}
To enable 3D geometric reasoning, we model target objects as cylinders $C = (r_m, h_m)$ with radius $r_m$ and height $h_m$ and assume that targets move on the ground plane ($z = 0$).
This recovery is applied to each $\mathbf{b}_d$ using a default cylinder with nominal height $h_m = 1.65$~m and radius $r_m = 0.3$~m following anthropometric conventions~\cite{pavlakos2019smplx}. 

\begin{figure}[htbp!]
    \centering
    \includegraphics[width=0.18\textwidth]{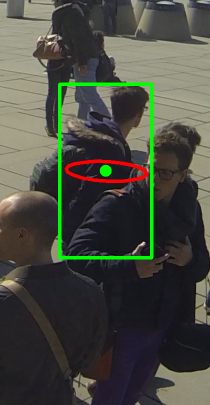}
    \includegraphics[width=0.18\textwidth]{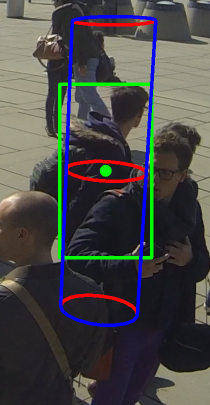}
    \includegraphics[width=0.18\textwidth]{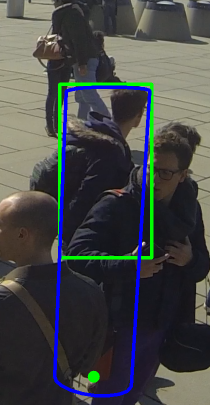}
    \caption{
        Full body bounding box and foot location recovered from an occluded detection:
        (left) projection of the cylinder model at the expected waist point $p_{\mathrm{waist}}$,
        (center) convex hull of the projected cylinder used to recover the full body,
        (right) adjusting the projection based on top-edge comparison to handle occlusions. 
    }
    \label{fig:SV3Ddemo}
\end{figure}

\begin{algorithm}[htbp!]
\caption{Recover 3D coordinates from bounding box}
\label{alg:recover3D}
\begin{algorithmic}[1]
\Require $\mathbf{b} = [u, v, w, h]$, cylinder model $C = (r_m, h_m)$, 
\newline 
camera projection matrix $P_c$ (per camera)
\Ensure recovered box $\mathbf{b}_{\mathrm{rec}}$, ground location $(x, y, 0)$, 
\newline 
visibility $v_{\mathrm{obj}}$, distance camera-object $d_{\mathrm{cam}}$
\State $p_{\mathrm{waist}} \leftarrow (u + w/2, v + h/2)$
\State Project $p_{\mathrm{waist}}$ to $z=h_m/2$ using $P_c$.
\State Project cylinder $C$ to the image using $P_c$.
\State Compute the convex hull of the projected cylinder (foot and head circles), 
\newline
take its axis-aligned bounding rectangle $\mathbf{b}_C$
\State $y_{\mathrm{diff}} \leftarrow v_C - v$ \hfill (compare top edges)
\If{$y_{\mathrm{diff}} < 0$}
    \State $y_{\mathrm{shift}} \leftarrow y_{\mathrm{diff}}$ \hfill (align top)
    \State $x_{\mathrm{shift}}$  proportional to the projective ``leaning''
\Else
    \State $y_{\mathrm{shift}} \leftarrow (h_C - h)/2$ \hfill
    \State $x_{\mathrm{shift}} \leftarrow 0$
\EndIf
\State $p_{\mathrm{waist}}^{\mathrm{adj}} \leftarrow (p_{\mathrm{waist}}^x - x_{\mathrm{shift}}, p_{\mathrm{waist}}^y - y_{\mathrm{shift}})$
\State Re-project and update $\mathbf{b}_{\mathrm{rec}}$ with adjusted waist
\State Use $P_c$ to back-project $p_{\mathrm{waist}}^{\mathrm{adj}}$ to world $z=0$ ($x, y, 0$)
\State Compute $v_{\mathrm{obj}}$ and $d_{\mathrm{cam}}$ using \cref{eq:visibility&dist2cam}
\State \Return $\mathbf{b}_{\mathrm{rec}}, (x, y, 0), v_{\mathrm{obj}}, d_{\mathrm{cam}}$
\end{algorithmic}
\end{algorithm}

We further assume that cameras are positioned at heights $z > h_m$ with angles $\theta < 90^\circ$ with respect to the $z$-axis to map each $\mathbf{b}_d$ to a location on the ground plane.
When these assumptions are met, most partial occlusions affect only the lower portion of the target.
\cref{fig:SV3Ddemo} illustrates the steps for full body recovery.
The waist point $(u, v)$ is back-projected to the world plane at $z = h_m/2$ using the camera projection matrix; 
the cylinder is placed there and projected back into the image to obtain a predicted silhouette (blue convex hull in \cref{fig:SV3Ddemo}). 
We then compare this silhouette to $\mathbf{b}_d$. 
If the projected model is taller than the detection, we treat the lower body as occluded and align the top edges;
otherwise we align the bottom edges to refine the foot location. 
Algorithm~\ref{alg:recover3D} summarizes the steps.

This procedure yields the recovered (full-body) bounding box of the projected cylinder ($\mathbf{b}_{\mathrm{rec}}$) and ground-plane foot location $(x, y)$.
From those, we derive the target \textit{visibility} $v_{\mathrm{obj}}$ and camera-target distance $d_{\mathrm{cam}}$ according to
\begin{align}
\label{eq:visibility&dist2cam}
v_{\mathrm{obj}} = \min\left(1, \frac{\mathrm{area}(\mathbf{b})}{\mathrm{area}(\mathbf{b}_{\mathrm{rec}})}\right),
\; \; 
d_{\mathrm{cam}} = \left\| \mathbf{c}_{\mathrm{cam}} - (x, y, h_m/2) \right\|_2,
\end{align}
where $\mathbf{c}_{\mathrm{cam}}$ is the camera location. These metrics are used for weighting associations, gating matches, and prioritizing targets in multi-view fusion stages.

\subsection{Data Association}
\begin{figure}[htbp!]
    \centering
    \vspace{-4mm}

\begin{tikzpicture}[
    state/.style={ellipse, draw=black, very thick, text=black, align=center, 
    minimum width=10mm, minimum height=8mm, font=\scriptsize\bfseries},
    arrow/.style={->, >=Stealth, thick, dashed, draw=black},
    label/.style={font=\scriptsize, text=black, align=center}
]


\node[state, fill=svblue!20]   (tentative)   at (0.0,0) {Tentative};
\node[state, fill=svblue!20]      (active)      at (3,0) {Active};
\node[state, fill=mvorange!20] (quasiactive) at (6,0) {Quasi-Active};
\node[state, fill=svblue!20]  (terminated)  at (1, -2) {Terminated};
\node[state, fill=svblue!20]    (inactive)    at (4, -2) {Inactive};

\draw[arrow] (tentative) to[out=120,in=60,looseness=3] node[above, label] {} (tentative);
\draw[arrow] (tentative) to[out=30,in=150,looseness=1] node[above, label] {} (active);
\draw[arrow] (tentative) to[out=250,in=150,looseness=1] node[left, label, xshift=-5pt] {} (terminated);

\draw[arrow] (active) to[out=120,in=60,looseness=3] node[above, label] {} (active);
\draw[arrow] (active) to[out=30,in=150,looseness=1] node[above, label] {} (quasiactive);
\draw[arrow] (active) to[out=250,in=135] node[left, label, xshift=-5pt] {} (inactive);

\draw[arrow] (quasiactive) to[out=120,in=60,looseness=3] node[above, label] {} (quasiactive);
\draw[arrow] (quasiactive) to[out=210,in=330] node[above, label] {} (active);
\draw[arrow] (quasiactive) to[out=270,in=0] node[right, label, xshift=8pt] {} (inactive);
\draw[arrow] (quasiactive) to[out=320,in=300,looseness=1.5] node[right, label, xshift=8pt] {} (terminated);

\draw[arrow] (inactive) to[out=240,in=300,looseness=3] node[below, label] {} (inactive);
\draw[arrow] (inactive) to[out=100,in=290] node[left, label, xshift=-5pt] {} (active);
\draw[arrow] (inactive) to[out=210,in=330] node[left, label, xshift=-5pt] {} (terminated);

\draw[arrow] ([xshift=-10mm]tentative.west) -- (tentative.west);
\draw[arrow] ([xshift=10mm]quasiactive.east) -- (quasiactive.east);

\end{tikzpicture}

\vspace{-15mm}

    \caption{
MV3DT track lifecycle and recovery logic.
Tracks begin as Tentative, are promoted to Active after a short probation with consistent matches, and fall back to Inactive for shadow tracking when detections are missed.
Quasi-Active denotes targets confirmed by peer cameras.
enabling multi-view continuity, while Terminated closes stale tracks.
    }
    \label{fig:statetransition}
\end{figure}
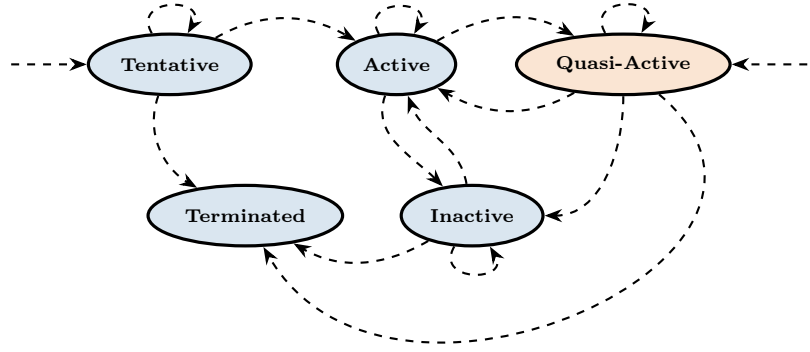

Data association modules ensure the consistency of target identities on multiple views and across time.
They depend on the target tracking state, which can be one of the following:
\begin{description}
    \item [Tentative:] on probation but not confirmed.
    \item [Active:] confirmed and visible.
    \item [Quasi-Active:] confirmed in other views but not visible.
    \item [Inactive:] confirmed but currently not visible.
    \item [Terminated:] no longer available.
\end{description}
The transitions between tracking states are shown in~\cref{fig:statetransition}.

\subsubsection{Single View Data Association}
ensures the consistency of target identities from frame to frame by matching detections $\mathbf{b}_d$ to target predictions $\mathbf{b}_t$. 
Data association is performed using intersection over union as an overlap similarity $S_{\mathrm{IoU}}$, size similarity $S_{\mathit{Size}}(d, t) = \min\left(\frac{\mathrm{area}(\mathbf{b}_d)}{\mathrm{area}(\mathbf{b}_t)}, \frac{\mathrm{area}(\mathbf{b}_t)}{\mathrm{area}(\mathbf{b}_d)}\right)$ (ratio of bounding box areas), and ReID similarity $S_{\mathrm{ReID}}$ for appearance.

\subsubsection{ReID Features for Appearance.}
A dedicated ReID module extracts a feature vector $\mathbf{f}_d$ from each detection crop using a ReID model as in~\cite{wojke2017deepsort} and maintains a gallery of such features $\{\mathbf{f}_{t,i}\}$ from each target's history. ReID similarity $S_{\mathrm{ReID}}$ is given by the maximum cosine distance between the detection feature and the elements of the gallery: $S_{\mathrm{ReID}}(d, t) = \max_i \mathbf{f}_d^\intercal \mathbf{f}_{t,i}$ (with $\|\mathbf{f}_d\| = 1$).

The overall similarity for \textbf{data association} is a convex combination:
\begin{equation}
    \label{eq:globalsimilarity}
    S_{\mathrm{Global}} = w_1 S_{\mathrm{ReID}} + w_2 S_{\mathrm{IoU}} + w_3 S_{\mathit{Size}},\qquad w_1 + w_2 + w_3 = 1.
\end{equation}
Similar to ByteTrack~\cite{zhang2022bytetrack}, our method performs greedy association at each phase, but it comprises three logically distinct stages:
\begin{itemize}
    \item \textbf{Stage 1}: High confidence detections are associated with confirmed \textbf{Active} and \textbf{Inactive} targets.
    \item \textbf{Stage 2}: \textbf{Active} targets unmatched in the first stage are associated with low confidence detections.
    \item \textbf{Stage 3}: Remaining high confidence detections are associated with \textbf{Tentative} targets. 
\end{itemize}

\subsubsection{Multi-View Association} \label{sec:MV3DT}
 shares data among cameras to maintain track continuity. 
From the multi-camera point of view, we define the ego-cam as the camera view under consideration, and peer-cams as cameras in the network whose FOVs overlap with the ego-cam's. 
Each ego-cam stores tracklet data from its peer-cams in a \textsc{PeerTargetDB} structure.

\paragraph{Multi-View Tracklet Matching.}
Central to multi-view association is \textbf{tracklet matching}: 
we compare ego and peer tracklets, each a sequence of $(f, x, y)$ (frame index and 3D feet positions). 
We search over a bounded time shift $\Delta t \in [-\Delta_{\max}, \Delta_{\max}]$ to account for small synchronization offsets between cameras. 
For each shift we collect frames where both tracklets are present, and if at least $n\geq n_{\min}$ common frames are found, we compute $S_T = 1/(1 + \sqrt{D^2/n})$, where $D^2 = \sum_i d_i^2$, is the sum of the Euclidean distances between the two feet positions at the $i$-th common frame. We take the maximum score over all shifts. 
Algorithm~\ref{alg:trackletMatching} summarizes the procedure.
Our multi-view target association mechanism allows for fully distributed ID propagation, which is executed simultaneously by each camera and consists of three stages.

\begin{algorithm}[htbp!]
\caption{Tracklet match score (ego vs.\ peer)}
\label{alg:trackletMatching}
\begin{algorithmic}[1]
\Require Tracklets $\mathcal{T}_1$, $\mathcal{T}_2$ (lists of $(f, x, y)$), time-shift range $\Delta_{\max}$,
\newline
minimum common frames $n_{\min}$
\Ensure Similarity score $S_{\mathrm{T}} \in [0,1]$
\State Build map: frame $f \mapsto$ index in $\mathcal{T}_2$
\State $S_{\mathrm{max}} \gets 0$
\For{$\Delta t = -\Delta_{\max}$ \textbf{to} $\Delta_{\max}$}
\State $n \gets 0$, $D^2 \gets 0$
\For{each $(f_1, x_1, y_1)$ in $\mathcal{T}_1$}
\State Look up $f_2 = f_1 + \Delta t$ in $\mathcal{T}_2$ to get $(x_2, y_2)$
\If{entry exists}
\State $D^2 \gets D^2 + (x_1{-}x_2)^2 + (y_1{-}y_2)^2$; $n \gets n + 1$
\EndIf
\EndFor
\If{$n \geq n_{\min}$}
\State $S \gets 1 / (1 + \sqrt{D^2/n})$; $S_{\mathrm{max}} \gets \max(S_{\mathrm{max}}, S)$
\EndIf
\EndFor
\State \Return $S_{\mathrm{max}}$
\end{algorithmic}
\end{algorithm}

\paragraph{Stage 1 -- Re-Association at ID Acquisition.} 
When a target becomes \textbf{Active}, we first try to re-associate it within the ego-cam (matching and recovering a shadow track) as described in~\cref{sec:targetmanagement}. 
If no match is found, we match the target trajectory against \textsc{PeerTargetDB} and adopt a peer ID if $S_T > \tau_{peer}$.
Since multiple cameras may simultaneously track the same target, the IDs are timestamped and preference is given to the oldest ID. 
This ensures ID convergence for the same target across cameras.
If both steps fail, the target receives a new ID.

\paragraph{Stage 2 -- Late Peer Re-Association.}
In certain scenarios, such as initialization or in the presence of non-negligible communication delays, tracks on different cameras may become \textbf{Active} nearly simultaneously.
To alleviate this problem, we allow tracklet matching to occur for a period $T_{rec}$ (in frames) after the target's activation.
During this period, the target is considered recently active.

\paragraph{Stage 3 -- ID Correction.}
When multiple targets with similar motion patterns are close to each other or when targets remain in the image border for a long period, their 3D location estimates can be inaccurate, potentially leading to incorrect associations.
This stage detects them by performing tracklet matching between associated targets. 
When any associated target no longer meets the matching criteria, i.e., $S_T < \tau_{peer}$, the algorithm checks for the age of the targets, and the older target keeps the ID.

\subsection{Target Management} \label{sec:targetmanagement}
Target management handles the lifecycle of tracked objects and their trajectories based on ego-cam and peer-cam data.

\subsubsection{Single-View Target Management.}
Following the core principles of SORT~\cite{bewley2016sort} and DeepSORT~\cite{wojke2017deepsort}, 
and as illustrated in \cref{fig:statetransition},
unmatched detections initialize new tracks in \textbf{Tentative} mode, and
after a probation age, the target becomes \textbf{Active} if it accumulates enough matches.
Targets without a current match transition to \textbf{Inactive}, which keeps track through misdetections and brief occlusions by updating the target locations using motion prediction, a strategy we call shadow tracking.

Since target motion estimates can drift during shadow tracking, recovering a target after occlusion based solely on its current position can be inaccurate.
Hence, we maintain a re-association database (\textsc{ReAssocDB}) that stores trajectory projections from targets in shadow tracking. 
When a track becomes \textbf{Active} after its probation period, its trajectory is compared with all \textbf{Inactive} trajectories in \textsc{ReAssocDB} to obtain $S_{\mathrm{T}}$. If a match occurs ($S_{\mathrm{T}} > \tau_{ego}$), the new track inherits the identity and history of the corresponding \textbf{Inactive} target. 

If an \textbf{Inactive} target is matched with a detection before reaching its maximum shadow tracking age, it transitions back to \textbf{Active}.
Otherwise, it is terminated and its trajectory removed from \textsc{ReAssocDB}.

\subsubsection{Multi-View Target Management.}
The following mechanisms use peer data to manage the lifecycle of the targets.

When no measurements are available in the ego-cam, we can use peer-cam observations to keep tracking the target. 
We introduce the \textbf{quasi-active} state for those cases.
In quasi-active tracking, the target state is updated using only peer-cam measurements.
The target returns to active if matched with a detection or is terminated if not matched and peer measurements are no longer available. 

To better address the early birth of targets, our method also initiates the association process based on peer-cam observations.
When a new target is partially detected (i.e., it is becoming unoccluded or is at the edges of the FOV) or far from the camera, its detections can have very low confidence.
If such detections align with a confident peer target, they probably represent an actual target.
Thus, MV3DT starts a new track in these early detection cases.

Similar to the early tracker instantiation, when targets are fully occluded before the ego-cam can see them for the first time, MV3DT can ``see-through'' the occlusions by instantiating a new track for such targets.
A see-through target is initialized in quasi-active mode, as long as a peer-cam provides a confident state and the target location projects onto the ego-cam FOV.

\subsection{Multi-View Measurement Fusion.}
Motion estimation is performed by a Kalman filter~\cite{kalman1960filter} to maintain the state (feet location and constant velocity) of each target and predict their locations for \textbf{data association} using ego-cam measurements.
Our system also fuses measurements whenever a target is visible in different camera views. 
However, factors such as low visibility or a long distance from the camera degrade measurement quality, sometimes so much so that a peer measurement should be treated as an outlier. 
We therefore apply the following criteria before fusing a peer measurement with the ego-cam estimate: 
i) it projects onto the ego-cam FOV; 
ii) its visibility (from the monocular detection module) is greater than a minimum threshold;
iii) its distance from the ego-cam target prediction is less than a threshold;
iv) the peer target visibility is greater than the ego target visibility; and 
v) the distance from the target to the peer camera is smaller than its distance to the ego-cam.
Criteria (i)--(iii) ensure the measurement is geometrically and qualitatively acceptable; (iv) and (v) ensure that fusing it actually improves over the ego estimate.
Measurements meeting these criteria are fed into the state estimator jointly with the ego-cam measurements to update the target states. 
Each Kalman filter iteration runs one prediction step and multiple correction steps, one for each measurement.

\subsection{Inter-camera Communication} \label{sec:communications}
The communication module is based on a publish/subscribe paradigm, in which each camera publishes its data on a dedicated topic and other cameras selectively subscribe to these topics to receive the information they require. 

\begin{figure}[t]
    \centering
\begin{tikzpicture}[
    field/.style={draw=black, rotate=45, thick, minimum height=6mm, minimum width=25mm, font=\footnotesize\bfseries, align=center, inner sep=2pt},
    group label/.style={font=\footnotesize, text=darkgray},
    node distance=0.5pt
]
    \node[field, fill=lightgray!35] (f1) at (0,0) {frame};
    \node[field, fill=lightgray!35] (f2) at (10mm,0) {camID};
    \node[field, fill=lightgray!35] (f3) at (20mm,0) {targetID};
    \node[field, fill=lightgray!35] (f4) at (30mm,0){targetID Ts};
    \node[group label] at (15mm,-15mm) {all};

    \node[field, fill=svblue!25] (f5) at (40mm,0) {targetAge};
    \node[field, fill=svblue!25] (f6) at (50mm,0) {state};
    \node[field, fill=svblue!25] (f7) at (60mm,0) {stateTime};
    \node[field, fill=svblue!25] (f8) at (70mm,0) {visibility};
    \node[field, fill=svblue!25] (f9) at (80mm,0) {camDist};
    \node[field, fill=svblue!25] (f10) at (90mm,0) {tracklet};
    \node[group label] at (65mm,-15mm) {tracklet, stateUpdate};

    \node[field, fill=mvorange!30] (f11) at (100mm,0) {prevID};
    \node[group label] at (100mm,-15mm) {adoptedID};
\end{tikzpicture}

\vspace{-3mm}
    \caption{Message fields: all message types  include \texttt{frame}, \texttt{camID}, \texttt{targetID}, and \texttt{targetID Ts} (timestamp). \textbf{tracklet} and \textbf{stateUpdate} also carry \texttt{targetAge}, \texttt{state}, \texttt{stateTime}, \texttt{visibility}, and \texttt{camDist}; \textbf{tracklet} further includes the \texttt{tracklet} payload, while \textbf{adoptedID} adds only \texttt{prevID} (the ID replaced by \texttt{targetID}).}
    \label{fig:messagefields}
\end{figure}
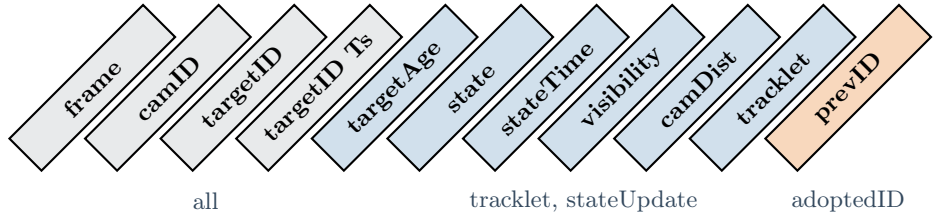

\paragraph{Message Types.}
The protocol defines three message types (Fig.~\ref{fig:messagefields}):
\textbf{tracklet}, sent when a target becomes active and used to create entries in the \textsc{PeerTargetDB};
\textbf{stateUpdate}, sent after each Kalman update to refresh existing entries; and
\textbf{adoptedID}, sent when an ID changes due to late peer re-association or ID correction to update the corresponding entry.

The communication and data processing overhead in MV3DT is dependent on a configurable subscription graph. 
For large-scale deployments, a naive, fully-connected graph (``all-to-all'' configuration) would flood every camera with irrelevant data. 
This creates high network usage and CPU overhead. 
To prevent this, cameras only subscribe to its peers, i.e., other cameras with overlapping FOVs.

MV3DT's modular design allows for different communication technologies and protocols.
We currently support an in-memory module for single-computer deployments, using shared memory for low-latency message passing, and an MQTT-based module~\cite{mqttMQTTStandard} for distributed setups, using Eclipse Mosquitto~\cite{eclipse2026mosquitto} as the message broker.

\section{Experiments and Results} \label{sec:experiments}

We evaluate our approach on three benchmark datasets, comparing \textbf{3D tracking accuracy} and \textbf{scalability} against state-of-the-art methods. Our experiments demonstrate that MV3DT achieves competitive accuracy while offering superior scalability for large-scale deployments.

We use NVIDIA DeepStream~\cite{nvidia2026deepstream} for video ingestion and generic people detectors from NVIDIA TAO~\cite{nvidia2026tao}. We denote as \textbf{PNT} the transformer-based detector PeopleNet Transformer~1.1~\cite{nvidia2026peoplenet-transformer} used in the tracking accuracy assessment and as \textbf{PN3} the lightweight detector PeopleNet 2.6.3~\cite{nvidia2026peoplenet} used for better time performance in the scalability evaluation.

\subsection{Datasets}
\textbf{WILDTRACK}~\cite{chavdarova2018wildtrack} is a widely used multi-view pedestrian tracking benchmark featuring 7 synchronized cameras with overlapping fields of view covering an outdoor plaza. 
The dataset provides video sequences for 2,000 frames at 10 FPS with 400 annotated frames at 2 FPS. 
Camera calibrations are also provided.
\textbf{SCOUT}~\cite{engilberge2025one} is a recent larger scale dataset with 25 calibrated cameras capturing real outdoor pedestrian traffic over long sequences. 
It provides 12,000 frames at 10 FPS covering a 450-meter path, with ground truth annotations for 8 cameras.
\textbf{AI City 2024 Warehouse Synthetic Dataset}~\cite{wang2024aicitychallenge} is one of the AI City Challenge scenes representing a 100-camera simulation of a large-scale warehouse environment for 6 minutes. 
It features dense camera coverage, and realistic occlusion patterns. 
The dataset is used for stress-test scalability and real-time performance analysis.

\subsection{Tracking Accuracy on WILDTRACK}

\cref{tab:wildtrack} compares MV3DT against recent state-of-the-art multi-view tracking methods on WILDTRACK.
MV3DT achieves 96.5\% IDF1, 93.1\% MOTA, and 94.6\% MOTP at 27~FPS, tying the best reported IDF1 while achieving the highest MOTP compared to the state-of-the-art methods. 
Although UMPN~\cite{engilberge2025one} and MVTrajecter~\cite{yamane2025mvtrajecter} achieve slightly higher MOTA, they rely on learned, scene-specific models. 
MV3DT trades marginal MOTA for deployability, scalability, and real-time throughput.

\begin{table}
\centering
\fontsize{9}{9}\selectfont
\caption{Results on the WILDTRACK test set: last 10\% of the 7 sequences.}
\setlength{\tabcolsep}{6pt}
\begin{tabular}{lcccc}
\toprule
Method                              & IDF1$\uparrow$ & MOTA$\uparrow$ & MOTP$\uparrow$ & FPS$\uparrow$ \\
\midrule
  EarlyBird~\cite{chavdarova2018wildtrack} & 92.3 & 89.5 & 86.6 & -- \\
  MVTr~\cite{yamane2025mvtrajecter}        & 93.1 & 92.3 & 92.7 & -- \\
  BEV-SUSHI~\cite{teepe2024lifting}        & 93.4 & 87.5 & 94.3 & -- \\
  MVFlow~\cite{engilberge2023mvflow}       & 93.5 & 91.3 &  --  & -- \\
  TrackTacular~\cite{teepe2024lifting}     & 95.3 & 91.8 & 85.4 & -- \\
  UMPN~\cite{engilberge2025one}            & 96.3 & 93.9 & 86.9 & 2  \\
  MVTrajecter~\cite{yamane2025mvtrajecter} & \bf 96.5 & \bf 94.3 & 93.0 & -- \\
  \bf MV3DT (Ours) w/ PNT                  & \bf 96.5 & 93.1 & \bf 94.6 & 27 \\ 
\bottomrule
\end{tabular}
\label{tab:wildtrack}
\vspace{-5mm}
\end{table}

\subsection{Tracking Accuracy and Performance on SCOUT}

\cref{tab:scout} presents results on the labeled subset of the SCOUT dataset (8 annotated cameras), using a 50\% train/test split as in~\cite{engilberge2025one}. 
MV3DT outperforms the baseline (UMPN+SP) by a margin of +14.7 IDF1 +25.9 MOTA and +19.3 MOTP percentage points when using PNT.
With the PN3 detector, MV3DT also shows scale-up capabilities: higher throughput and still better accuracy than UMPN.

\begin{table}
\centering
\fontsize{9}{9}\selectfont
\caption{Results on the SCOUT test set: last 50\% of the 8 annotated cameras.}
\setlength{\tabcolsep}{6pt}
\begin{tabular}{lcccc}
\toprule
Method & IDF1$\uparrow$ & MOTA$\uparrow$ & MOTP$\uparrow$ & FPS$\uparrow$\\
\midrule
ByteTrack MV~\cite{engilberge2025one} & 24.3 & 19.0 & 25.7 & -\\
UMPN~\cite{engilberge2025one}         & 26.9 & 24.1 & 26.1 & 2\\
UMPN + SP                             & 27.0 & 25.0 & 27.8 & -\\
\bf MV3DT (Ours) w/ PN3               & \bf 34.4 & \bf 44.3 & \bf 42.8 & \bf 55 \\
\bf MV3DT (Ours) w/ PNT               & \bf 41.7 & \bf 50.9 & \bf 47.1 & \bf 28 \\
\bottomrule
\end{tabular}
\label{tab:scout}
\vspace{-5mm}
\end{table}

\subsection{Ablation of the ID propagation protocol}

To quantify the contribution of each stage of our ID propagation protocol, we run an ablation on WILDTRACK by progressively enabling Stages 1--3 of multi-view association.
As shown in \cref{tab:ablation}, multi-view processing dramatically improves ID consistency over a single-view baseline (multi-view communication disabled). 
Adding Stage 1 (ID Acquisition) more than doubles both metrics, showing the benefit of cross-camera matching. 
Late Re-Association brings a particularly large gain on WILDTRACK, due to the short sequence length that allows for a significant proportion of nearly simultaneous activation during pipeline initialization.
Finally, ID Correction further reduces residual ID errors, reaching 96.5\% IDF1 and 93.1\% MOTA, matching the full-system WILDTRACK results in \cref{tab:wildtrack}.

\begin{table}[htbp!]
\centering
\fontsize{9}{9}\selectfont
\caption{Ablation of the three-stage ID propagation on WILDTRACK. Each row adds the indicated component. ``Single-view only'' disables multi-view interactions.}
\setlength{\tabcolsep}{6pt}
\begin{tabular}{lcc}
\toprule
Configuration & IDF1$\uparrow$ & MOTA$\uparrow$ \\
\midrule
Single-view only               & 25.0 & 15.9 \\
+ Stage 1 (ID Acquisition)     & 55.4 & 44.9 \\
+ Stage 2 (Late Re-Assoc.)     & 87.5 & 89.6 \\
+ Stage 3 (ID Correction)      & 96.5 & 93.1 \\
\bottomrule
\end{tabular}
\label{tab:ablation}
\vspace{-5mm}
\end{table}

\subsection{Large-Scale Deployment}

To validate the scalability of our method, we use two NVIDIA H100 NVL GPUs to process the AI City 2024 Warehouse Synthetic Dataset (50 streams per GPU).

\paragraph{Synchronized real-time performance:} 
Barrier-based frame synchronization limits each camera's input stream to at most one frame per $1/f_s$ period. In this scenario, MV3DT sustains 30 FPS across all 100 streams.
GPU utilization averages 95\% and the network bandwidth for inter-camera communication is 44~MB/s total (0.44 MB/s per camera), indicating efficient resource usage and demonstrating the lightweight nature of the communications.
The average inter-camera message latency, which is critical for real-time multi-view association, is 5 ms with the 95th percentile at 12 ms, well below the typical 33 ms real-time budget.

\paragraph{Asynchronous stress test:} Without frame synchronization, MV3DT achieves 44.5~FPS on 100 streams.
For comparison, a single-view baseline achieves 45.5~FPS, indicating the overhead of distributed ID propagation and fusion of only 2.2\%.
\cref{tab:stress} reports tracking accuracy with and without frame synchronization on the 100-camera warehouse scene.
We also provide the current leaderboard on the AI City Challenge, as a reference of expected accuracy. However, note that our accuracy evaluation is only for the specific warehouse scene. 

\begin{table}[htbp!]
\centering
\fontsize{9}{9}\selectfont
\caption{Tracking accuracy on the 100-camera AI City Warehouse Scene: synchronized vs.\ unsynchronized (stress test). Leaderboard results on the complete AICity'24 dataset are provided for reference. These methods do not report real-time performance but are unlikely to achieve real-time operation at such a scale.}
\setlength{\tabcolsep}{6pt}
\begin{tabular}{lcccc}
\toprule
Method / Setup & IDF1$\uparrow$ & HOTA $\uparrow$ & FPS$\uparrow$\\
\midrule
Single-View                      &  9.3 & 15.1 & 45.5 \\
Synchronized w/ PN3              & 79.5 & 68.6 & 30.0 \\
Unsynchronized w/ PN3            & 72.8 & 68.3 & 44.5 \\
(Un)synchronized w/ PNT          & 82.7 & 71.1 & 15.6 \\
\midrule
SJTU-Lenovo~\cite{AICity24Paper1} & -- & 67.2 & -- \\
Yachiyo~\cite{AICity24Paper32}    & -- & 71.9 & -- \\
BEV-SUSHI~\cite{wang2024bev}      & -- & 81.2 & -- \\
\bottomrule
\end{tabular}
\label{tab:stress}
\vspace{-5mm}
\end{table}

\section{Conclusions} \label{sec:conclusions}

MV3DT addresses a critical gap in multi-camera tracking: the need for real-time, scalable systems that operate without scene-specific training or retraining for different camera configurations. Unlike learned approaches, such as UMPN, BEV-SUSHI, and MVTrajecter, MV3DT is a zero-shot method that requires only camera calibrations. Most competing methods cannot be deployed to different camera configurations without retraining, effectively locking solutions to their training geometry.

Our experiments demonstrate that MV3DT achieves competitive accuracy on WILDTRACK, SCOUT, and the AI City 2024 warehouse scene, while providing superior scalability. 
On the 100-camera warehouse setup, MV3DT sustains 30~FPS with 5 ms average inter-camera latency and only 2.2\% communication overhead. On SCOUT (8 cameras), it reaches 28--55 FPS depending on the detector, compared to 2 FPS for UMPN. 
The ability to scale out by adding more nodes and cameras is where the architectural difference matters most. 
Most methods require all cameras to share a single GPU, so throughput and memory bounds prevent them from scaling beyond a handful of cameras. 
MV3DT, in contrast, is designed for distributed deployment: each camera stream runs in a separate process (machine or GPU), and nodes communicate only via lightweight messaging (e.g., MQTT). 
Scaling is thus limited by network bandwidth rather than GPU memory or compute. 
This positions MV3DT as uniquely suited for large-scale real-world deployments in warehouses, airports, and smart cities.

\bibliographystyle{splncs04}
\bibliography{ref}
\end{document}